# Enhance Multi-domain Sentiment Analysis of Review Texts through Prompting Strategies

Yajing Wang
*Hong Kong Baptist University,*
*BNU-HKBU United International College*
*Faculty of science and technology,*
*Department of computer science*
Hongkong, Zhuhai, China
wangyajing@uic.edu.cn

Zongwei Luo*
*BNU-UIC Institute of AI and Future Networks,*
*Artificial Intelligence and Data Science Research Hub*
*Beijing Normal University,*
*BNU-HKBU United International College*
Zhuhai, China
lzwqhk@outlook.com

*Abstract*—Large Language Models (LLMs) have made significant strides in both scientific research and practical applications. Existing studies have demonstrated the state-of-the-art (SOTA) performance of LLMs in various natural language processing tasks. However, the question of how to further enhance LLMs' performance in specific task using prompting strategies remains a pivotal concern. This paper explores the enhancement of LLMs' performance in sentiment analysis through the application of prompting strategies. We formulate the process of prompting for sentiment analysis tasks and introduce two novel strategies tailored for sentiment analysis: RolePlaying (RP) prompting and Chain-of-thought (CoT) prompting. Specifically, we also propose the RP-CoT prompting strategy which is a combination of RP prompting and CoT prompting. We conduct comparative experiments on three distinct domain datasets to evaluate the effectiveness of the proposed sentiment analysis strategies. The results demonstrate that the adoption of the proposed prompting strategies leads to a increasing enhancement in sentiment analysis accuracy. Further, the CoT prompting strategy exhibits a notable impact on implicit sentiment analysis, with the RP-CoT prompting strategy delivering the most superior performance among all strategies.

I. INTRODUCTION

Large Language Models (LLMs) are one of the core research points of neural natural language processing (NLP) in recent years. Academia and industry have produced a large number of research results around large language models: InstructGPT [1], LaMDA [2], ERNIE 3.0 Titan [3] etc. ChatGPT is a chat robot publicly released by OpenAI, an American artificial intelligence startup company, in November 2022. It creates content in the form of multiple rounds of dialogue, guided by user prompts. ChatGPT has the functions of theme creation, knowledge question answering and programming calculation in daily use situations. In the process of interacting with humans, it shows amazing performance by tracking the dialogue process, admitting mistakes and correcting and rejecting unreasonable requests. Impressive language skills.

Researches show that LLMs represented by chatGPT have reached or even surpassed the average human level in multiple tasks such as language comprehension, reasoning, and dialogue generation [4]. Existing studies have shown that ChatGPT has excellent performance on many English datasets, and even some evaluation indicators of the data set are comparable to the supervised fine-tuning model [5]. ChatGPT surpasses state-of-the-art (SOTA) performance under zero-shot in areas such as machine translation, sentiment analysis, and disinformation detection [6]. For example, Wang et al. [7] evaluate ChatGPT's ability to understand opinions, sentiments, and emotions in text. It compares ChatGPT with other models, showing the fabulous performance of ChatGPT in different sentiment analysis settings. Ji et al. [8] evaluate ChatGPT's performance in text-based personality recognition, particularly its zero-shot chain-of-thought prompting and optimized level-oriented prompting strategies, uncovering its capabilities, fairness concerns, and impact on downstream tasks. Gao et al. [9] present a comprehensive evaluation of ChatGPT's causal reasoning capabilities, revealing its strengths as a causal interpreter while uncovering its limitations in causal reasoning, particularly concerning causal hallucination, sensitivity to prompt phrasing, and performance variations based on event characteristics in sentences.

However, the mentioned studies have primarily assessed ChatGPT from a macroscopic perspective, comparing it with other models for specific tasks, without delving into the nuanced exploration of how different prompting strategies may impact ChatGPT's performance. This arouses an important research question: how to enhance ChatGPT's task-specific performance through prompting techniques? Sentiment analysis is a classic NLP task. The model needs to determine the emotional categories contained in it according to the input context [10]. In this study, we use sentiment analysis as the representative task to explore how to enhance the performance of tasks through prompting strategies. In order to improve the reliability of this study, we conduct experiments using datasets from multiple domains.

This paper try to explore effective prompting strategies to further enhance the performance of ChatGPT in sentiment analysis tasks. Specifically, we propose a prompting strategy named RolePlaying-CoT (RP-CoT) for the sentiment analysis task, which integrate role playing strategy into CoT prompting. Comparative experiments on multiple-domain datasets are conducted to identify the optimal prompting strategy tailored for sentiment analysis tasks using ChatGPT.

II. RELATED WORK

*A. Sentiment Analysis with LLMs*

Recent studies have shown that large-scale pre-trained language models, such as BERT and GPT, have achieved

* Corresponding author.
This work was supported by Research Grants Council of Guangdong Universities Special Key Project (Project No. 2021ZDZX3021), and Guangdong Higher Education Upgrading Plan (2021-2025) of "Rushing to the Top, Making Up Shortcomings and Strengthening Special Features" with UIC research grant UICR0400052-21CTL.

SOTA performance on various NLP tasks, including sentiment analysis [11-12]. However, the capability boundaries of these models are still not well understood, and there is a need for further evaluation and analysis of their performance on specific tasks [13].

One such model that has drawn attention in the research community and the public is ChatGPT, a variant of the GPT model that has been fine-tuned on conversational data [14]. While ChatGPT has shown impressive performance on various conversational tasks, its performance on sentiment analysis tasks is still not well understood. To address this gap, Wang et al. [7] conducted a preliminary evaluation of ChatGPT's performance on five representative sentiment analysis tasks and 18 benchmark datasets. The evaluation involved four different settings, including standard evaluation, polarity shift evaluation, open-domain evaluation, and sentiment inference evaluation, and compared ChatGPT with fine-tuned BERT-based models [15] and corresponding SOTA models on each task for reference.

The results of the evaluation showed that ChatGPT exhibits impressive zero-shot performance in sentiment classification tasks and can rival fine-tuned BERT, although it falls slightly behind the domain-specific fully-supervised SOTA models [7]. The study also included human evaluation and qualitative case studies, which provided further insights into ChatGPT's sentiment analysis capabilities. These findings suggest that ChatGPT has potential as a universal sentiment analyzer, although further research is needed to fully understand its capability boundaries and limitations.

*B. Prompt Learning Methods*

Prompt learning is a recent approach in NLP that aims to improve the performance of language models on specific tasks by providing them with task-specific prompts or instructions. This approach has gained significant attention in the NLP community due to its ability to achieve SOTA results on a wide range of tasks, including text classification, question answering, and language generation [16].

Several methods have been proposed for prompt learning, including template-based approaches [17-19], where prompts are manually designed based on the task requirements, and gradient-based approaches [20-21], where prompts are learned through gradient descent optimization. Other methods include prompt tuning [22-24], where prompts are fine-tuned on a specific task, and prompt programming [25], where prompts are generated using a programming language. Despite the success of prompt learning, there are still several challenges that need to be addressed, such as prompt design and selection, prompt generalization, and prompt overfitting. Nevertheless, prompt learning is a promising direction for improving the performance of language models on specific tasks and has the potential to revolutionize the field of NLP.

Shinn et al. [26] introduces a sophisticated formulation that allows the model to iterate and externalize thoughts before generating a final response. This approach demonstrates enhanced performance in downstream language tasks, particularly suited for intricate reasoning requirements. Conversely, Fulford and Ng [27] contribute to this discourse by initiating an online course that imparts general principles for crafting prompts, emphasizing clarity and specificity of instructions, as well as granting the model time for contemplation. Emergent investigations in the contemporary landscape [28-29] have revealed that LLMs possess the capacity to simulate the role of an anticipated agent when provided with comprehensive and intricate delineations. Xu et al. [30] introduces Expert-Prompting, utilizing In-Context Learning to create specialized prompts, enhancing the answering quality of large language models. Through this strategy, a new chat assistant, ExpertLLaMA, is trained, demonstrating improved performance compared to existing open-source models.

Chain-of-thought (CoT) [31] prompting process is a recently developed prompting method that encourages large language models to explain their reasoning processes. The main idea of the chain of thought is to explain the reasoning process in examples by showing a small number of exapmles to the large language model, and the large language model will also show the reasoning process when answering the prompt. Interpretations of this kind of reasoning tend to lead to more accurate results. Fig.1. [31] shows an example of CoT prompting enable LLMs solving a math problem correctly. We can see from Fig.1. that the correct answer cannot be made with the standard model. But if given the model some ideas about problem-solving, just like a math test will write out the problem-solving process and finally get the answer. CoT prompting is to decompose a multi-step reasoning problem into many intermediate steps and assign them to more calculation amount, generate more tokens, and then splice these answers together for solution. You can see in Fig.1 that using CoT prompting can get the correct mathematical answer.

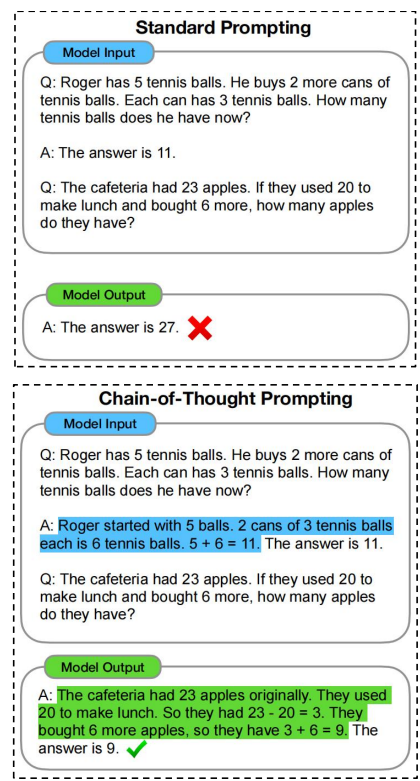

Fig. 1. An example of CoT prompting enable LLMs solving a math problem correctly.

### III. PROMPTING STRATEGIES FOR SENTIMENT ANALYSIS

In this chapter we will introduce three prompting strategies for the sentiment analysis task of LLMs. They are inspired by the above mentioned prompting learning methods, and the third prompting strategy is a combination of the former two prompting strategies.

## A. Role-playing Prompting for Sentiment Analysis

To illustrate the prompting strategies, we formulate the prompting process as follows:

$$y = LLM(x, i) \quad (1)$$

As shown in (1), suppose we give the review text x and instruction i, the LLM would generate an answer y.

The idea of role-playing is easy to understand. In short, identifying role and add this role to the prompt. Since our objective task is specific, the straightest way to identify role is identifying by human. It is intuitive that the identifying role could be an expert of sentiment analysis. It is worth noting that when the task conduced by LLMs is not specific, for example, there are numerous random conversations as instructions, at this time, automatic method such as In-Context Learning [16] is suggested to be used to accurately distinguish roles. After identifying the role, the role-playing (RP) prompting process can be formulated as follows:

$$y = LLM(RP, x, i) \quad (2)$$

The difference of (1) and (2) is that there is one more RP. RP means the role playing in the prompt. According to the role-playing prompting strategy, a fixed description (RP) can be added at the beginning of the instruction:

> You are an expert of sentiment analysis, {instruction}...

Further, we can make the RP more specific by adding the domain for the role. For example, if we want the LLM conduct sentiment analysis of the movie review texts, the detailed RP can be added, shown as follows:

> You are an expert of sentiment analysis in movie review texts domain, {instruction}...

We will test the effectiveness of the role-playing prompting strategy in the subsequent experimental part.

## B. CoT Prompting for Sentiment Analysis

In the realm of social media commentaries, a substantial presence of implicit sentiment is discernible. Unlike explicit sentiment, implicit sentiment content within these texts lacks overt lexical markers indicative of explicit emotional biases. Implicit sentiment analysis refers to the task of identifying the sentiment expressed in text without the use of explicit sentiment indicators, such as positive or negative words [32].

The complexity of implicit sentiment analysis surpasses that of explicit sentiment analysis due to the necessity for the model to grasp contextual intricacies and underlying semantic nuances of the text. Although LLMs have achieved SOTA performance in overall sentiment analysis [7], their outcomes pertaining to implicit sentiment, generated under conventional prompts, may not invariably be accurate (refer Vanilla Prompting example in the Appendix).

To rectify this, prompting strategies are imperative to steer these models towards generating accurate implicit sentiment assessments. Subsequently, this discourse will elucidate the utilization of CoT prompting for guiding Language Models in the domain of sentiment analysis.

We find that the implicit sentiment is difficult to be recognized in the long sentence but easier to be identified given the specific aspect. We draw on the idea of "divide and conquer" to construct CoT prompting for the sentiment analysis and divide the prompting into three steps. Referring the research [33], the prompting process can be formulated as follows:

$$y_1 = LLM(x, i_1)$$

$$y_2 = LLM(y_1, i_2)$$

$$y = LLM(y_1, y_2, i) \quad (3)$$

In (3), x means the sentence input, $i_1$, $i_2$, i refer to the instruction in every shot, and $y_1$, $y_2$ are the intermediate outcomes generated by the LLM. The three steps of the CoT prompting strategy are shown as follows:

*1) Let the LLM distinguish what is the aspect described of the subjective for the given text:*

> Given the review text x, what is the specific aspect described of the subjective?

*2) Classify the sentiment polarity for the identified aspect and obtain the rationale:*

> What is the implicit opinion towards the mentioned aspect, and why?

*3) Based on the former two context knowledge, classify the sentiment polarity for the text:*

> Based on above, {instruction}...

With regard to the process, the subsequent two steps leverage the logical progression established in the preceding step. This entails the decomposition of the overarching problem into smaller constituent queries, wherein the solution obtained in the previous step serves as foundational background knowledge for addressing the subsequent query.

## C. RolePlaying-CoT Prompting for Sentiment Analysis

The RolePlaying prompting technique serves to bestow upon the LLM a precise persona aligned with the task, thereby enhancing the congruence between the LLM's outputs and the intended human objectives. Conversely, the CoT prompting approach involves the dissection of complex inquiries, injecting logical cues to guide the model's reasoning process and facilitate the generation of accurate responses. Both of these prompting strategies contribute to enhancing the effectiveness of sentiment analysis through prompts. The amalgamation of these two strategies is readily conceivable. This section delineates the combination of RolePlaying prompting and CoT prompting within the context of sentiment analysis tasks, hereby referred to as the RolePlaying-CoT (RP-CoT) prompting strategy.

ChatGPT employs a Transformer-based autoregressive language model, which integrates a self-attention mechanism. Additionally, ChatGPT incorporates Long Short-Term Memory (LSTM). These mechanism facilitates the machine's comprehension and encapsulation of contextual cues within dialogues, enabling the model accurately capture the contextual nuances of conversations, consequently enhancing its proficiency in maintaining a seamless context for continuous dialogue interactions. Therefore, to combine the two strategies, we just need to add the RP prompting strategy at the beginning of the CoT promoting process, for the

sentiment analysis task. The RP-CoT prompting strategy for sentiment analysis can be formulated as (4):

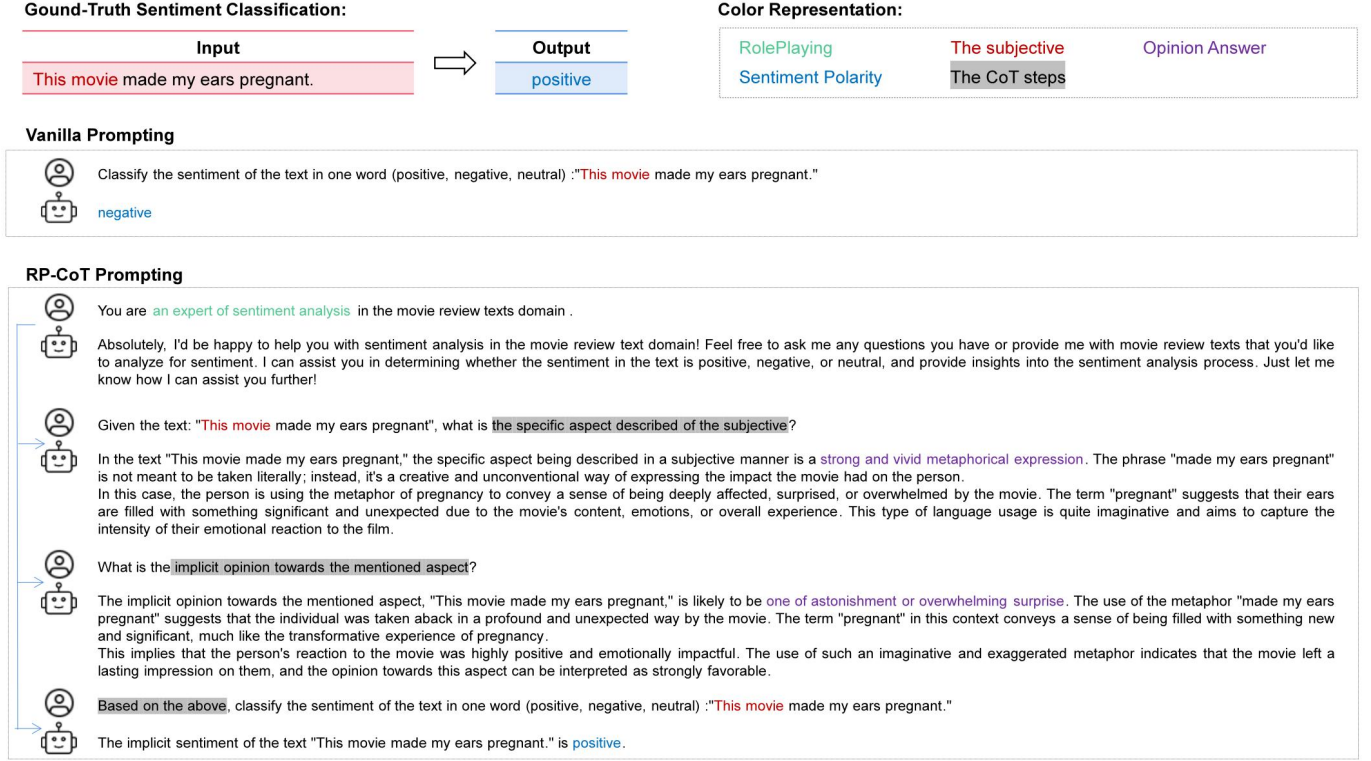

Fig. 2. An example for illustrating the RP-CoT prompting strategy.

$$y_1 = \text{LLM}(RP)$$

$$y_2 = \text{LLM}(y_1, x, i_1)$$

$$y_3 = \text{LLM}(y_1, y_2, i_2)$$

$$y = \text{LLM}(y_1, y_2, y_3, i) \quad (4)$$

The difference between (3) and (4) is that there are one more initiation to assign the RP to the LLM in (4) compared with (3). Overall, there are four steps for the RP-CoT prompting strategy for sentiment analysis:

*1) Assign the RP to the LLM:*

> You are an expert of sentiment analysis in ... domain, please follow the instructions step by step.

*2) Let the LLM distinguish what is the aspect described of the subjective for the given text:*

> Given the review text x, what is the specific aspect described of the subjective?

*3) Classify the sentiment polarity for the identified aspect and obtain the rationale:*

> What is the implicit opinion towards the mentioned aspect, and why?

*4) Based on the former two context knowledge, classify the sentiment polarity for the text:*

---
[1] http://ai.stanford.edu/~amaas/data/sentiment/
[2] https://www.kaggle.com/datasets/sbhatti/financial-sentiment-analysis
[3] https://www.kaggle.com/datasets/bittlingmayer/amazonreviews

> Based on above, {instruction}...

In the first step, we use the Zero Shot Chain of Thought (Zero-shot-CoT) to help the LLM form a thought chain. The consequent steps are the same as the CoT prompting strategy for sentiment analysis. In this way, the LLM can "think better" under the role background. We take an example to illustrate our RP-CoT prompting strategy, shown in Fig.2.

## IV. EXPERIMENTS AND RESULTS

In this chapter, we will employ diverse prompting strategies to conduct sentiment analysis on review texts from various domains using ChatGPT. This endeavor aims to validate the performance of distinct prompting strategies across multiple domains for sentiment analysis purposes.

### A. Datasets

We choose three classical review text datasets with labels for sentiment analysis in three different domains (movie, finance and shopping).

- IMDB dataset[1]. IMDB dataset is a movie review dataset for binary sentiment classification (positive and negative) containing 50K movie reviews for natural language processing or Text analytics.

- FiQA, Financial PhraseBank datasets[2]. This financial review dataset is a combination of two datasets (FiQA, Financial PhraseBank). It provides financial sentences with sentiment labels (positive, negative and neutral), intended for advancing financial sentiment analysis research [34].

- Amazon Reviews[3]. This dataset consists of a few million Amazon customer reviews and star ratings (1-5 stars) for learning how to train fastText for sentiment analysis. Here we preprocess 1- and 2-star reviews as negative label, and 4- and 5-star reviews as positive label.

*B. Baselines*

To evaluate the performance of prompting strategies proposed in section III, we will compare four prompting strategies for sentiment analysis of review texts.

- Vanilla prompting strategy. The vanilla prompting strategy serves as the plainest prompting, where no extra prompting strategies added to the instructions.

- RP prompting strategy. The RP prompting strategy involves assigning a role to the LLM before providing instructions, allowing the LLM to function as a domain expert through proxy means.

- CoT prompting strategy. The CoT prompting strategy adheres to the concept of the thought chain, breaking down a question into several sub-questions and guiding the LLM to learn the logical reasoning required to arrive at the correct answers.

- RP-CoT prompting strategy. The RP-CoT prompting strategy is a combination of RP prompting strategy and CoT prompting strategy. It assigns a role to the LLM during the Zero-Shot phase and subsequently guides the LLM through the process of CoT-based problem-solving.

*C. Setups*

We conducted sentiment analysis using ChatGPT on the three aforementioned datasets, employing the four prompting strategies delineated above. The experiments conduct the model GPT-3.5, operating exclusively in a zero-shot setting, with a fixed temperature parameter set to zero, in order to mitigate result variability. It's worth noting that the three datasets vary in label categories; the movie and shopping datasets classify review texts into positive and negative polarities, whereas the financial dataset comprises three polarities, including a neutral category. Consequently, when providing instructions, we also account for the disparity in polarity count. Our experiments were conducted on a server equipped with 4 NVIDIA A100 GPUs.

*D. Evaluations*

The most crucial metric for assessing sentiment analysis performance is accuracy. We employ the "accuracy_score" function from the "sklearn.metrics" library in Python to directly compute the accuracy score. A higher accuracy score signifies better sentiment analysis performance. Table 1 summarizes the accuracy results for sentiment analysis of ChatGPT across three-domain datasets using four different prompting strategies.

TABLE I. ACCURACY RESULTS FOR SENTIMENT ANALYSIS

| Accuracy | Movie | Finance | Shopping |
|---|---|---|---|
| Vanilla prompting | 0.918 | 0.696 | 0.923 |
| RP prompting | 0.920 | 0.721 | 0.943 |
| CoT prompting | 0.923 | 0.834 | 0.944 |
| RP-CoT prompting | 0.925 | 0.835 | 0.944 |

We also visualize the results in a line graph for the purpose of facilitating comparative observations and analysis, shown is Fig.3.

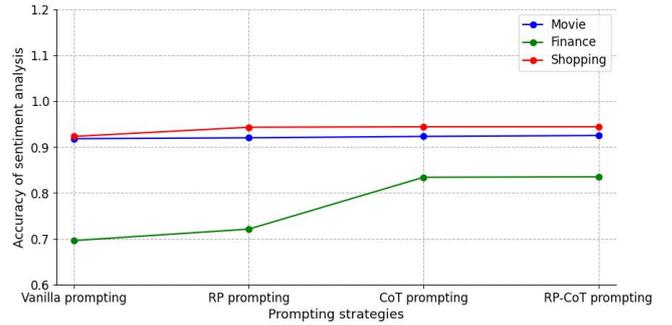

Fig. 3. Accuracy of sentiment analysis using different prompting strategies in three-domain review texts.

It can be seen in Fig.3 that across datasets from three domains, there is a consistent upward trend in the accuracy of sentiment analysis when employing the vanilla prompting strategy, RP prompting strategy, CoT prompting strategy, and RP-CoT prompting strategy. This trend shows the effectiveness for the prompting strategies, especially RP-CoT prompting strategy, having the best performance.

We also notice the accuracy of sentiment analysis for the movie and shopping datasets is higher and more stable compared with the accuracy of sentiment analysis for the finance dataset. One possible reason is the label in movie and shopping datasets is two-polarity (positive and negative), whereas the label in the finance dataset is three-polarity (positive, negative and neutral). Generally, the greater the number of labels in sentiment classification, the more challenging the task becomes. As demonstrated in Fig.2, it is also noted that for implicit sentiment, classifiers often struggle to capture the underlying meanings within sentences, leading to misclassifications. Similarly, within the finance dataset, there may exist numerous instances of implicit sentiment that conventional classifiers may fail to recognize, resulting in lower accuracy compared to the other two datasets. However, following the adoption of the CoT prompting strategy, a noticeable improvement in accuracy is observed, thereby validating the capability of the CoT prompting strategy in identifying implicit sentiment.

Table I presents specific data regarding the accuracy of sentiment analysis achieved with different prompting strategies. It is evident that the adoption of prompting strategies enhances the accuracy of sentiment analysis, with a notable performance improvement observed, particularly in the context of implicit sentiment classification.

To further substantiate the reason of the enhancement in sentiment analysis accuracy attributed to the RP-CoT prompting strategy, we generate confusion matrices for the sentiment analysis outcomes of the three-domain datasets under Vanilla prompting and RP-CoT prompting conditions. These matrices are depicted in Fig.4.

In each confusion matrix presented in Fig.4, distinct colors are employed to denote specific classification outcomes. Colors along the diagonal from the upper-left to the lower-right, tending towards yellow (indicating higher values), represent better classification performance, while colors in other regions tending towards purple (indicating lower values) signify poorer classification performance.

We can observe that the confusion matrices for the movie and shopping datasets show little variation between Vanilla prompting and RP-CoT prompting conditions. However, the confusion matrices for the finance dataset exhibit dissimilarity. Specifically, in the two confusion matrices of the finance dataset, under the RP-CoT prompting strategy, the diagonals are closer to yellow compared to Vanilla prompting. Upon closer examination, it becomes evident that there are more instances of misclassifications where negative sentiments are predicted as positive and neutral sentiments are predicted as negative. This is apparent from the lower-left region of the matrices, which is less aligned with purple compared to other areas. This also validates our earlier conjecture that the finance dataset indeed contains a significant amount of implicit sentiment, which can, however, be effectively addressed through the RP-CoT prompting strategy.

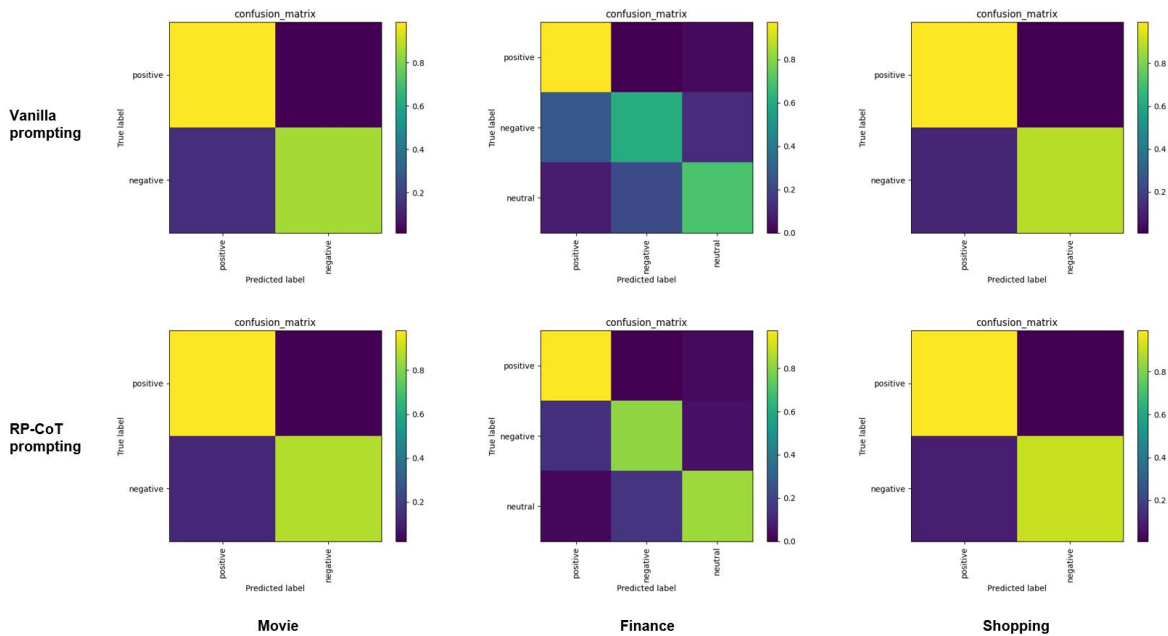

Fig. 4. Confusion matrices of sentiment analysis in three-domain datasets with Vanilla prompting and RP-CoT prompting, respectively. Each row refers to the confusion matrices using different prompting strategies (Vanilla and RP-CoT). Each column corresponds to the confusion matrices in different domain data: Movie, Finance and Shopping. Every subgraph shows the specific confusion matrix for the accuracy of labels and predicted classifications.

## V. CONCLUSIONS

This paper proposes three prompting strategies for sentiment analysis, RolePlaying (RP) prompting strategy, Chain-of-thought (CoT) strategy and RP-CoT prompting strategy, to enhance the sentiment analysis of review texts in multiple domains. The experiments conducted on three distinct domain datasets demonstrate the effectiveness of these strategies. The RP strategy leverages the role background of the reviewer to improve the sentiment analysis accuracy, while the CoT strategy helps LLMs form a thought chain to "think better" and capture the implicit sentiment of the review text. The RP-CoT prompting strategy which is a combination of both strategies further improves the performance of sentiment analysis. The proposed strategies have the potential to be applied to other domains and various tasks. We aspire for our work to contribute to the field of prompt engineering and make a meaningful impact on the advancement of NLP tasks.

## APPENDIX

An example using Vanilla prompting to classify the sentiment of a movie review with ChatGPT:

> **Y:** Classify the sentiment of the text in one word (positive, negative, neutral) :"This movie made my ears pregnant."
>
> **ChatGPT:** negative

An example using RP-CoT prompting to classify the sentiment of a movie review with ChatGPT:

> **Y:** You are an expert of sentiment analysis in the movie review texts domain
>
> **ChatGPT:** Absolutely, I'd be happy to help you with sentiment analysis in the movie review text domain! Feel free to ask me any questions you have or provide me with movie review texts that you'd like to analyze for sentiment. I can assist you in determining whether the sentiment in the text is positive, negative, or neutral, and provide insights into the sentiment analysis process. Just let me know how I can assist you further!
>
> **Y:** Given the text: "This movie made my ears pregnant", what is the specific aspect described of the subjective?
>
> **ChatGPT:** The phrase "This movie made my ears pregnant" seems to be used in a figurative and humorous way to describe how impactful or immersive the movie was. The word "pregnant" here is likely being used metaphorically to convey that the movie was so captivating and engaging that it filled the listener's ears with vivid and imaginative content, similar to how pregnancy involves nurturing and growth. In other words, the specific aspect being described here is the intense and creative impact the movie had on the person's auditory experience.
>
> **Y:** What is the implicit opinion towards the mentioned aspect?
>
> **ChatGPT:** The implicit opinion towards the mentioned aspect, which is the intense and creative impact of the movie on the person's auditory experience, is likely positive and appreciative. The use of the phrase "made my ears pregnant" is a playful and exaggerated way of expressing how much the person enjoyed and was immersed in the movie. This kind of colorful language usually indicates a strong positive reaction and a sense of being deeply affected by the movie's audiovisual elements.
>
> **Y:** Based on the above, classify the implicit sentiment of the text in one word (positive, negative, neutral) :"This movie made my ears pregnant."
>
> **ChatGPT:** positive